%% file: main.tex
\PassOptionsToPackage{dvipsnames,table}{xcolor}
\documentclass[10pt,twocolumn,letterpaper]{article}
\usepackage{float}
 \usepackage[pagenumbers]{iccv} 


\definecolor{iccvblue}{rgb}{0.21,0.49,0.74}
\usepackage[pagebackref,breaklinks,colorlinks,allcolors=iccvblue]{hyperref}
\usepackage{booktabs}
\usepackage[table]{xcolor}

\usepackage{placeins}

\newcommand{\method}{Zero-P-to-3}
\def\paperID{6017} 
\def\confName{ICCV}
\def\confYear{2025}

\title{Zero-P-to-3: Zero-Shot Partial-View Images to 3D Object}
\author{%
Yuxuan Lin\textsuperscript{1} \quad Ruihang Chu\textsuperscript{1} \quad Zhenyu Chen\textsuperscript{2} \quad Xiao Tang\textsuperscript{2} \quad Lei Ke\textsuperscript{1} \quad Haoling Li\textsuperscript{1} \\  \quad Yingji Zhong\textsuperscript{3} \quad Zhihao Li\textsuperscript{2} \quad Shiyong Liu\textsuperscript{2} \quad Xiaofei Wu\textsuperscript{2} \quad Jianzhuang Liu\textsuperscript{2} \quad Yujiu Yang\textsuperscript{1}\\[1ex]
\textsuperscript{1}Tsinghua University\\
\textsuperscript{2}Huawei Technologies Co., Ltd.\\
\textsuperscript{3}The Hong Kong University of Science and Technology
}

\begin{document}
\twocolumn [ 
{\renewcommand\twocolumn[1][]{#1}%
\maketitle
\begin{center}
    \centering
\vspace{-20pt}
\includegraphics[width=0.99\linewidth]{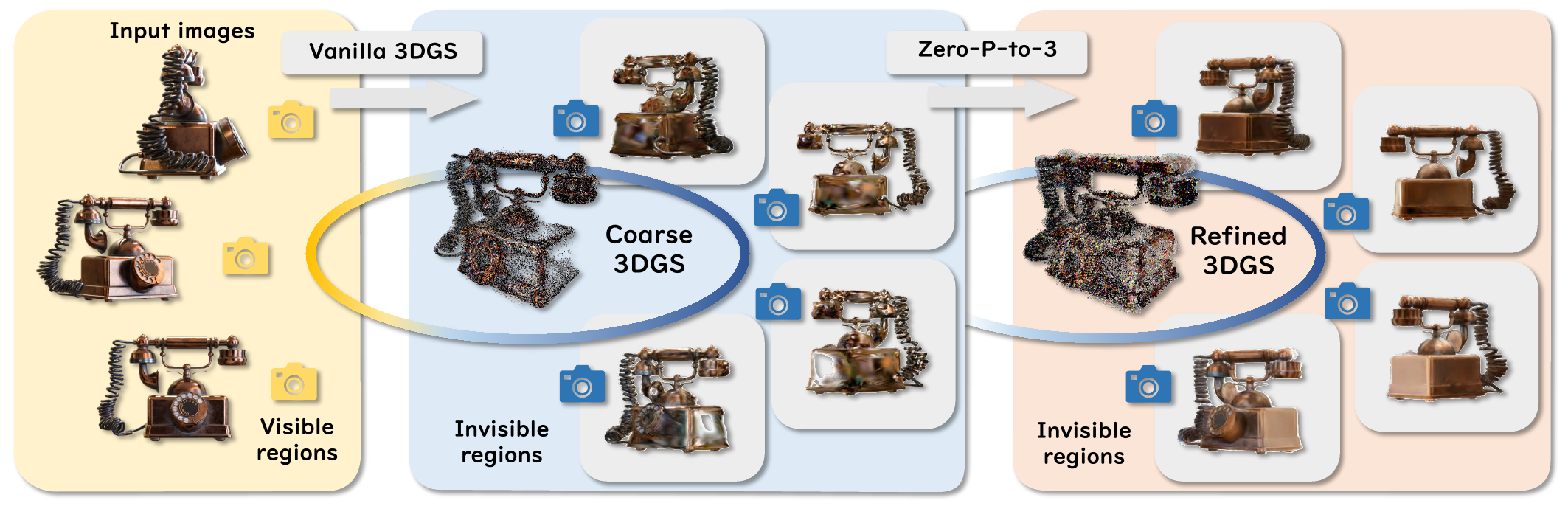}
\captionof{figure}{\method~reconstructs 3D objects where visible regions are confined to a limited field-of-view. We first reconstruct a coarse 3D Gaussian Splatting (3DGS) model from partial observations, then iteratively refine it using multiple priors to obtain the final 3DGS model.}
\label{fig:teaser}
\end{center}}]

\vspace{-7pt}
\input{0_abstract}    
\input{1_intro}
\input{2_related}

\input{3_methods}
\input{4_experiments}

\input{5_conclusion}

{
    \small
    \bibliographystyle{ieeenat_fullname}
    \bibliography{main}
}


\end{document}

%% file: 0_abstract.tex
\begin{abstract}
Generative 3D reconstruction shows strong potential in incomplete observations. While sparse-view and single-image reconstruction are well-researched, partial observation remains underexplored. In this context, dense views are accessible only from a specific angular range, with other perspectives remaining inaccessible. This task presents two main challenges:
(i) limited View Range: observations confined to a narrow angular scope prevent effective traditional interpolation techniques that require evenly distributed perspectives.
(ii) inconsistent Generation: views created for invisible regions often lack coherence with both visible regions and each other, compromising reconstruction consistency.
To address these challenges, we propose \method, a novel training-free approach that integrates the local dense observations and multi-source priors for reconstruction. Our method introduces a fusion-based strategy to effectively align these priors in DDIM sampling, thereby generating multi-view consistent images to supervise invisible views. We further design an iterative refinement strategy, which uses the geometric structures of the object to enhance reconstruction quality. Extensive experiments on multiple datasets show the superiority of our method over SOTAs, especially in invisible regions. 
\end{abstract}

%% file: 1_intro.tex
\vspace{-20pt}
\section{Introduction}
\label{sec:intro}
\vspace{10pt}
The field of 3D object reconstruction has witnessed significant evolution in recent years.
This task aims to reconstruct the geometry and texture of objects from available observations.
As complete observations are not always accessible due to occlusions or constraints in camera positioning, generative models leveraging prior knowledge have emerged as promising solutions. While existing approaches~\cite{li2024era3d,wu2024unique3d,long2024wonder3d,liu2023syncdreamer,wynn2023diffusionerf,yang2024gaussianobject,wu2024reconfusion,yu2024lm}, especially with diffusion models~\cite{rombach2022highresolutionimagesynthesislatent}, have made considerable progress, they mainly focus on reconstruction from a single image~\cite{li2024era3d,wu2024unique3d,long2024wonder3d,liu2023syncdreamer} or well-spaced sparse views~\cite{wynn2023diffusionerf,yang2024gaussianobject,wu2024reconfusion,yu2024lm,charatan2024pixelsplat3dgaussiansplats}.
Yet, partial observation scenarios are more common in daily life, where dense object observations are available from certain views but inaccessible from other angles.
For instance, in cluttered indoor environments, objects can be partially occluded by furniture or walls, making it difficult to capture all of their views.

3D reconstruction from a restricted range of observation angles presents unique characteristics and challenges.
Compared to single-image reconstruction, this setting provides dense observations over certain regions, offering more local information that has the potential to achieve high-quality reconstruction results. However, effectively combining these partial views to infer better structures remains underexplored. It differs from sparse-view reconstructions as well. Generally, these sparse views are often evenly distributed around the object and provide broader coverage~\cite{yang2024gaussianobject}, but partial observations confined to a limited range may leave substantial portions of the object completely unseen. As a result, traditional interpolation methods~\cite{niemeyer2022regnerf,wynn2023diffusionerf,charatan2024pixelsplat3dgaussiansplats,zhong2024cvt,yang2024gaussianobject} used in sparse-view reconstruction are not applicable.

Current generative reconstruction approaches may face critical challenges in partial observation scenarios. Most multi-view diffusion methods~\cite{long2024wonder3d,shi2023mvdream,li2024era3d,liu2023syncdreamer,liu2024one,wu2024unique3d} only take a single image as input to generate a fixed set of novel views. These generated views often exhibit inconsistencies both with the ground truth and among each other, preventing reliable supervision for reconstruction. Even advanced approaches that support multi-image inputs, such as LEAP~\cite{jiang2023leapliberatesparseview3d} and EscherNet~\cite{kong2024eschernetgenerativemodelscalable}, continue to struggle with coherent generation quality. This challenge arises because the 3D features corresponding to multiple inputs are not perfectly aligned, leading to inconsistencies in the generated content. Similarly, methods like Trellis~\cite{xiang2024structured3dlatentsscalable} that generate 3D content directly from latent 3D representations cannot guarantee consistency between the generated content and visible information, as the 3D latent space often fails to preserve the detailed information present in the input images. When applied to partial observation settings, these generation paradigms lack effective mechanisms to collectively utilize dense information from visible image sets. In contrast, neural reconstruction repair methods~\cite{yu2024lm,yang2024gaussianobject,wynn2023diffusionerf,wu2024reconfusion} operate by refining existing Gaussian representations. While these methods can effectively utilize information from visible regions, they leverage local information from other sparse views to enhance image features. Unlike the multi-view diffusion approaches, these repair mechanisms rely heavily on surrounding views for effective feature propagation, making them suitable when observations are well-distributed around 360$^{\circ}$, but less effective in partial observation settings where large regions remain completely invisible.

To address these challenges, we introduce a novel \textbf{Zero}-shot approach to transfer \textbf{P}artial-view image observations \textbf{to} reconstructed \textbf{3}D object, termed as \method. It is a training-free framework that integrates multiple priors and conditions for effective 3D reconstruction from partial observation. As illustrated in~Fig.~\ref{fig:teaser},
\method~begins by constructing a coarse 3D Gaussian Splatting model (3DGS)~\cite{kerbl20233d} using the available dense observations from views within a specific angular range. Building upon this initial reconstruction, we propose a fusion-based strategy that aligns various priors and conditions in the diffusion-based inference stage. Specifically, our method integrates information from multiple viewing angles, rendering results, multi-view diffusion priors, and image restoration priors into each denoising timestep. This fusion process generates a batch of high-quality, consistent images in novel views as the supervision, guiding the subsequent reconstruction process.

To further enhance the reconstruction quality, we design an iterative refinement framework to utilize rotated view sampling. Given that our observations cover a local region, we propose to rotate supervision views around these observed areas, thereby maximizing view coverage to provide dense supervision signals. This design efficiently leverages information from visible regions and enhances multi-view consistency by utilizing geometric structures and weak textures of the object during the rotation process, finally mitigating reconstruction artifacts.

Our \method~ presents significant superiority for partial observation reconstruction. By leveraging multiple priors and supervision from rotated views, we preserve fine details in observed regions while maintaining consistency across novel viewpoints. Unlike methods that require additional training on extra datasets, our \method~ provides a training-free solution adaptable to diverse scenarios. Experimental evaluations on objects with front-facing observations demonstrate that \method~ achieves significantly higher consistency with original observations compared to existing approaches. Our method effectively handles challenging cases involving complex geometry and occlusions. Our main contributions are as follows:
\begin{itemize}
    \item We investigate a practical yet challenging problem of 3D reconstruction from partial observations, and propose a novel training-free framework as a solution.
    \item We integrate multi-source priors with a rotated view refinement strategy to leverage local observations, enhancing the accuracy and detail of reconstruction. 
    \item Extensive experiments show our method's superiority in reconstruction quality and multi-view consistency.
\end{itemize}

%% file: 2_related.tex
\section{Reated Work}
\label{sec:realted}

\noindent \textbf{3D Sparse Reconstruction.} Traditional 3D sparse reconstruction methods primarily focus on scenarios with sparse but well-distributed viewpoints. RegNeRF~\cite{niemeyer2022regnerf} introduces color-based regularization terms, while DS-NeRF~\cite{deng2022depth} leverages depth supervision for robust reconstruction under sparse views. LEAP~\cite{jiang2023leapliberatesparseview3d} further addresses sparse-view 3D modeling by eliminating reliance on precise camera poses. CVT-xRF~\cite{zhong2024cvt} and Binocular-Guided 3D Gaussian Splatting~\cite{han2024binocularguided3dgaussiansplatting} advance this direction by modeling pixel relationships and view consistency through 3D Gaussian point cloud representations. Recent learning-based approaches show remarkable efficiency, with methods like LRM~\cite{hong2023lrm}, LGM~\cite{tang2025lgm}, and GRM~\cite{xu2024grm} leveraging large-scale models for efficient single-image or sparse-view reconstruction using transformer architectures. Generative models such as EscherNet~\cite{kong2024eschernetgenerativemodelscalable} explore scalable view synthesis through diffusion-based frameworks, while Structured3D~\cite{xiang2024structured3dlatentsscalable} introduces structured latent representations for versatile 3D generation. Similarly, Triplane Meets Gaussian Splatting~\cite{zou2024triplane}, Splatter Image~\cite{szymanowicz2024splatter}, and InstantMesh~\cite{xu2024instantmesh} focus on improving reconstruction speed while maintaining quality through various optimization techniques. Unlike these methods that rely on learned priors or training, our approach operates in a training-free manner while effectively handling partial dense observations.

\noindent \textbf{Diffusion-Based Generation and Reconstruction.} The integration of diffusion models has revolutionized 3D content generation and reconstruction. DreamFusion~\cite{poole2022dreamfusion} pioneers text-to-3D generation using 2D diffusion models, with Magic3D~\cite{lin2023magic3d} and Magic123~\cite{qian2023magic123} improving efficiency and quality. SJC~\cite{wang2023score} proposes an alternative optimization approach to Score Distillation Sampling, offering improved stability in knowledge transfer from 2D to 3D.  Zero123++~\cite{shi2023zero123++} and SyncDreamer~\cite{liu2023syncdreamer} focus on generating consistent multi-view images from a single view. For reconstruction, ReconFusion~\cite{wu2024reconfusion} and DiffusionERF~\cite{wynn2023diffusionerf} enhance 3D reconstruction using 2D image priors from diffusion models. GaussianObject~\cite{yang2024gaussianobject} achieves high-quality reconstruction from extremely limited views.LM-Gaussian~\cite{yu2024lm} introduces large-scale visual model priors for sparse-view reconstruction with robust initialization. Multi-view diffusion approaches like ViewCrafter~\cite{yu2024viewcrafter}, DimVis~\cite{di2024dimvis}, ReconX~\cite{liu2024reconx}, and ViewDiff~\cite{hollein2024viewdiff} demonstrate consistent multi-view image generation for 3D reconstruction. Recent approaches including MVDream~\cite{shi2023mvdream}, Wonder3D~\cite{long2024wonder3d}, SV3D~\cite{voleti2025sv3d}, and Era3D~\cite{li2024era3d} further improves generation quality and efficiency. While these methods typically reconstruct or generate from a single image or images covering the entire scene, whereas our approach uniquely combines diffusion priors with geometric constraints to handle partial dense observations.

%% file: 3_methods.tex
\section{Method}
\label{sec:methods}

\subsection{Task Definition}
\begin{figure}[t]
    \centering
    \includegraphics[width=1\linewidth]{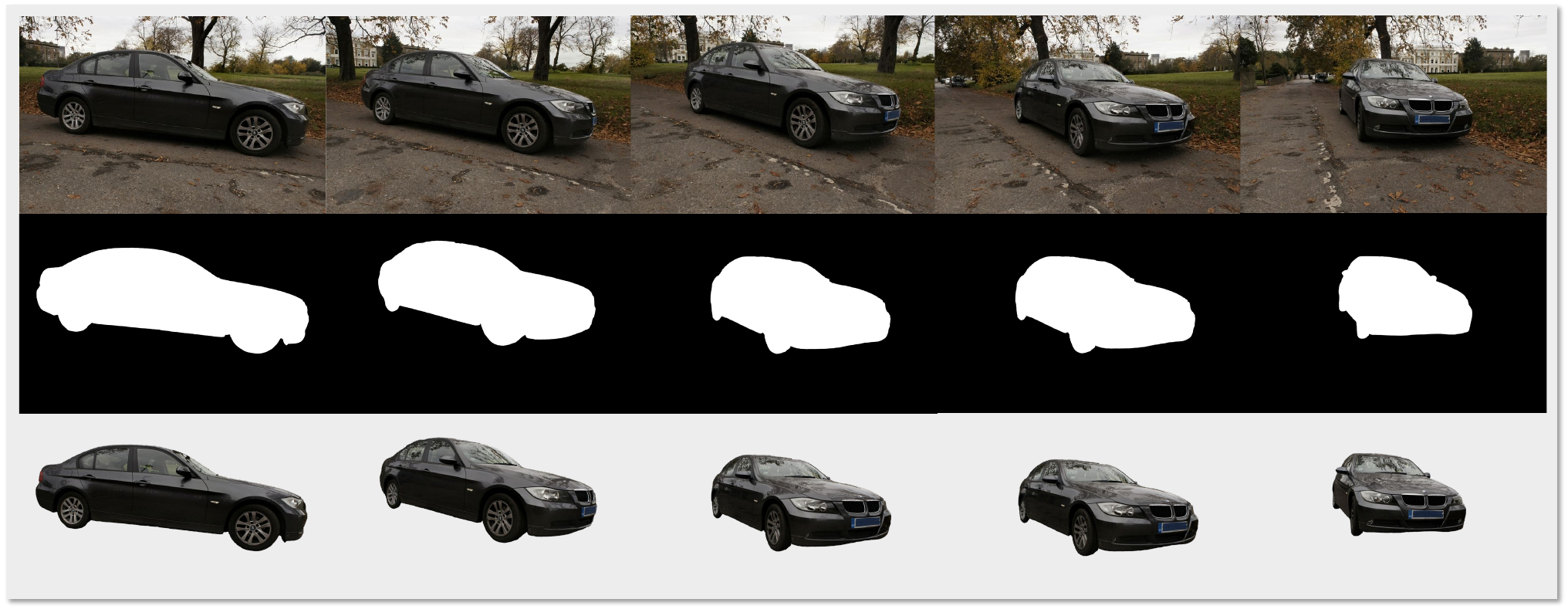}
    \caption{Visualizations of input images (top), masks (middle), and processed images (bottom) from the RefNeRF~\cite{verbin2022ref} dataset.}
    \vspace{-10pt} 
    \label{inputimages}
\end{figure}

Our work addresses the challenge of novel view synthesis from spatially constrained observations. Given a set of $N$ images $\mathbf{X}_\text{ref} = \{x_i\}_{i=1}^N$ captured from a limited spatial region (e.g., approximately $90^{\circ}$), we aim to construct a high-quality 3D Gaussian Splatting (3DGS) model $\mathcal{G}$ capable of generating photorealistic renderings from arbitrary viewpoints. As shown in~Fig.~\ref{inputimages}, these images are taken within a segment of a sphere, with each input image accompanied by its corresponding object mask $m_i$ generated by SAM, collectively represented as $\mathbf{M}_\text{ref} = \{m_i\}_{i=1}^N$.

\begin{figure*}[t]
    \centering
    \includegraphics[width=1\linewidth]{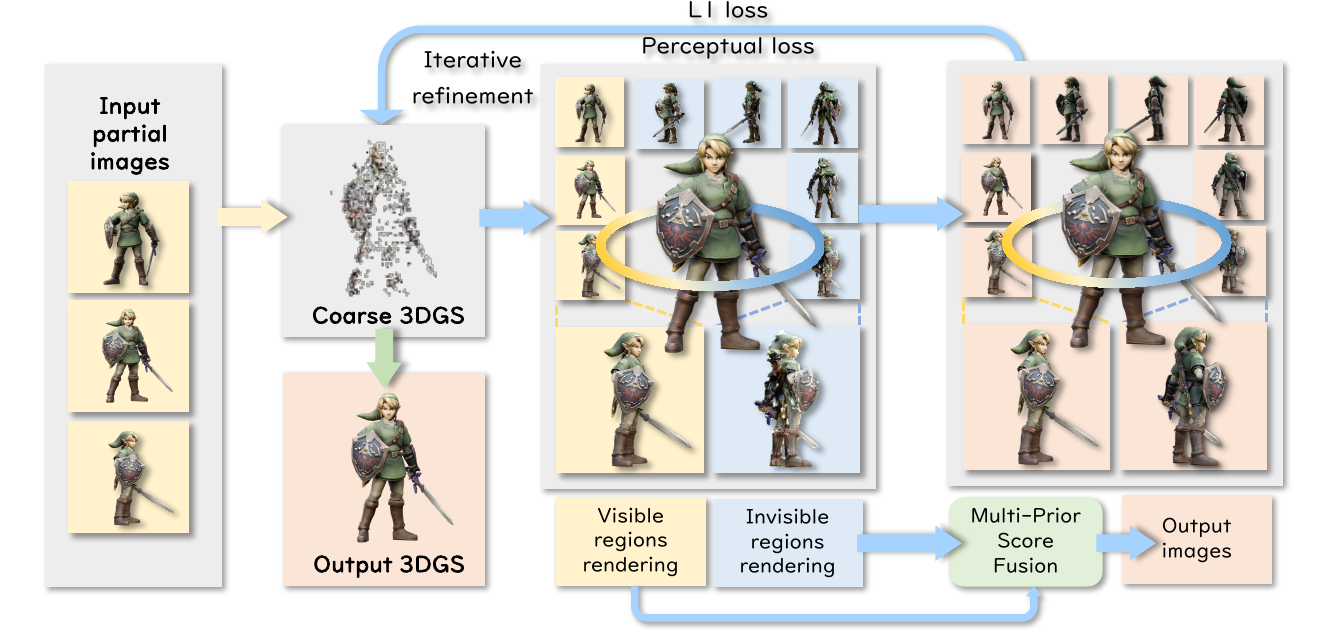}
    \caption{Overview of \method~for 3D reconstruction from partial observations. Starting with input partial images, the system first constructs a coarse 3D Gaussian Splatting (3DGS) model, then renders both visible and invisible regions. These renderings are enhanced through Multi-Prior Score Fusion to generate high-quality output images that serve as supervision for refining the 3DGS model. Through iterative refinement using L1 and perceptual losses, the system produces a comprehensive 3D representation that effectively bridges the gap between limited observations and complete object reconstruction.}
    \label{pipeline}
\end{figure*}

Our aim is to learn a function that generates new views based on arbitrary camera poses $\pi$, denoted as $x = \mathcal{G}(\pi | \{x_i, \pi_i, m_i\}_{i=1}^N)$. This challenging task requires addressing several key issues: establishing accurate geometric foundations from partial observations, generating high-quality and consistent images for invisible regions, and ensuring globally consistent optimization.  Our approach, illustrated in~Fig.~\ref{pipeline}, consists of three key stages: (1) coarse 3DGS initialization from partial observations, (2) multi-prior score fusion for view-consistent image generation, and (3) iterative refinement through rotated view sampling.
 \noindent

\subsection{Coarse Gaussians Initialization}
To initialize the 3DGS model, we first estimate the camera intrinsics \(\mathbf{K}_{\text{ref}}\) and extrinsics \(\Pi_{\text{ref}} = \{\pi_i\}_{i=1}^N\) for the input images \(\mathbf{X}_{\text{ref}}\) using Structure From Motion (SFM) techniques~\cite{schonberger2016structure}. Leveraging these parameters, we construct an initial coarse 3DGS model \(\mathcal{G}_0 = \{g_j\}_{j=1}^M\), where each Gaussian \(g_j\) is parameterized by a 3D position \(\mathbf{p}_j\), a covariance matrix \(\mathbf{\Sigma}_j\) defining its spatial extent, an opacity \(\alpha_j\), and spherical harmonics coefficients \(\mathbf{Y}_j\) for view-dependent color. The initialization is optimized by minimizing the photometric reconstruction error over visible regions across all input images:
\vspace{-5pt}
\begin{equation}
\vspace{-5pt}
\mathcal{G}_0 = \arg \min_{\mathcal{G}} \sum_{i=1}^N \sum_{\mathbf{p}_j \in \mathcal{V}_i} \left\| \mathcal{R}(\mathbf{p}_j, \mathbf{\Sigma}_j, \alpha_j, \mathbf{Y}_j, \pi_i) - x_i(\mathbf{p}_j) \right\|^2,
\end{equation}
where \(\mathcal{V}_i\) represents the set of 3D points visible in image \(x_i\), and \(\mathcal{R}(\cdot)\) denotes the volumetric rendering function of the 3DGS model. To enhance robustness, we incorporate the object masks \(m_i\) to focus optimization on the foreground object, reducing interference from background noise. While \(\mathcal{G}_0\) effectively captures the coarse geometry and appearance of visible areas, it typically exhibits incomplete details and inconsistencies in occluded or unseen regions, which necessitates a subsequent refinement stage.

\subsection{Multi-Prior Score Fusion}

To synthesize consistent, high-quality images for novel views, particularly in invisible regions, we propose a diffusion-based strategy termed Multi-Prior Score Fusion. Diffusion models generate images by iteratively estimating the score function \(\nabla_x \log p(x_t)\), progressively denoising from pure noise (\(t=1\)) to a clean image (\(t=0\)). Our goal is to synthesize a target view \( x_t \) conditioned on a set of reference views \( \{V_i\}_{i=1}^N \), extending multi-view diffusion (MVD) from a single-input to a multi-input framework. A simple approach to ensure consistency across inferences from different reference views is to average the individual view-conditioned scores:
\vspace{-5pt}
\begin{equation}
\vspace{-5pt}
    \nabla_x \log p(x_t | \{V_i\}_{i=1}^N) \approx \frac{1}{N} \sum_{i=1}^N \nabla_x \log p(x_t | V_i),
    \label{simple_fusion}
\end{equation}
where \(x_t\) is the target view at diffusion step \(t\), \(\{V_i\}_{i=1}^N\) is the set of \(N\) reference views, and \(\nabla_x \log p(x_t | \cdot)\) denotes the log-probability gradient of \(x_t\) conditioned on the respective views. This approximation combines gradient information from each reference view, guiding the synthesis of \( x_t \) toward a solution consistent with all inputs. By averaging, the method aligns inferences by assuming each view contributes equally to a unified representation of the target.

However, this approach has limitations. It assumes that each view contributes equally to the target view, though this varies significantly with the angle between them. More critically, it implies conditional independence among reference views, expressed as \( p(x_t | \{V_i\}) \propto \prod_{i=1}^N p(x_t | \{V_i\}_{i=1}^N) \), since the averaged score in Eq.~\ref{simple_fusion} reflects a product form when summing log-probabilities. In practice, such independence rarely holds due to correlated scene details across views, resulting in artifacts like blurriness in unobserved regions (see Fig.~\ref{fig:analysis}). We address this limitation through robust geometric and texture regularization, overcoming the theoretical flaws of the multi-view fusion approximation.

\subsubsection{Multi-View Fusion}
In our implementation, we use a standard multi-view diffusion (MVD) framework:
\vspace{-5pt}
\begin{equation}
\label{eq:mvd}
\vspace{-5pt}
\varepsilon_{\text{MVD}}(x_t) = \sum_{i=1}^N w_i \cdot \varepsilon_{\text{MVD}}(x_t | V_i),
\end{equation}
where \(\varepsilon_{\text{MVD}}(x_t | V_i)\) represents the noise predicted by a diffusion model conditioned on view \(V_i\), and \(w_i\) denotes the weight. For the weight configuration, frontal views are assigned a weight of 2, back views receive 1.5, and other viewpoints are allocated a weight of 1. In spite of this, the core challenge stems from correlations between views that invalidate the independence assumption, requiring a more substantive solution.

\subsubsection{Frequency-Decomposed Regularization}
\begin{figure}[t]
    \centering
    \includegraphics[width=1\linewidth]{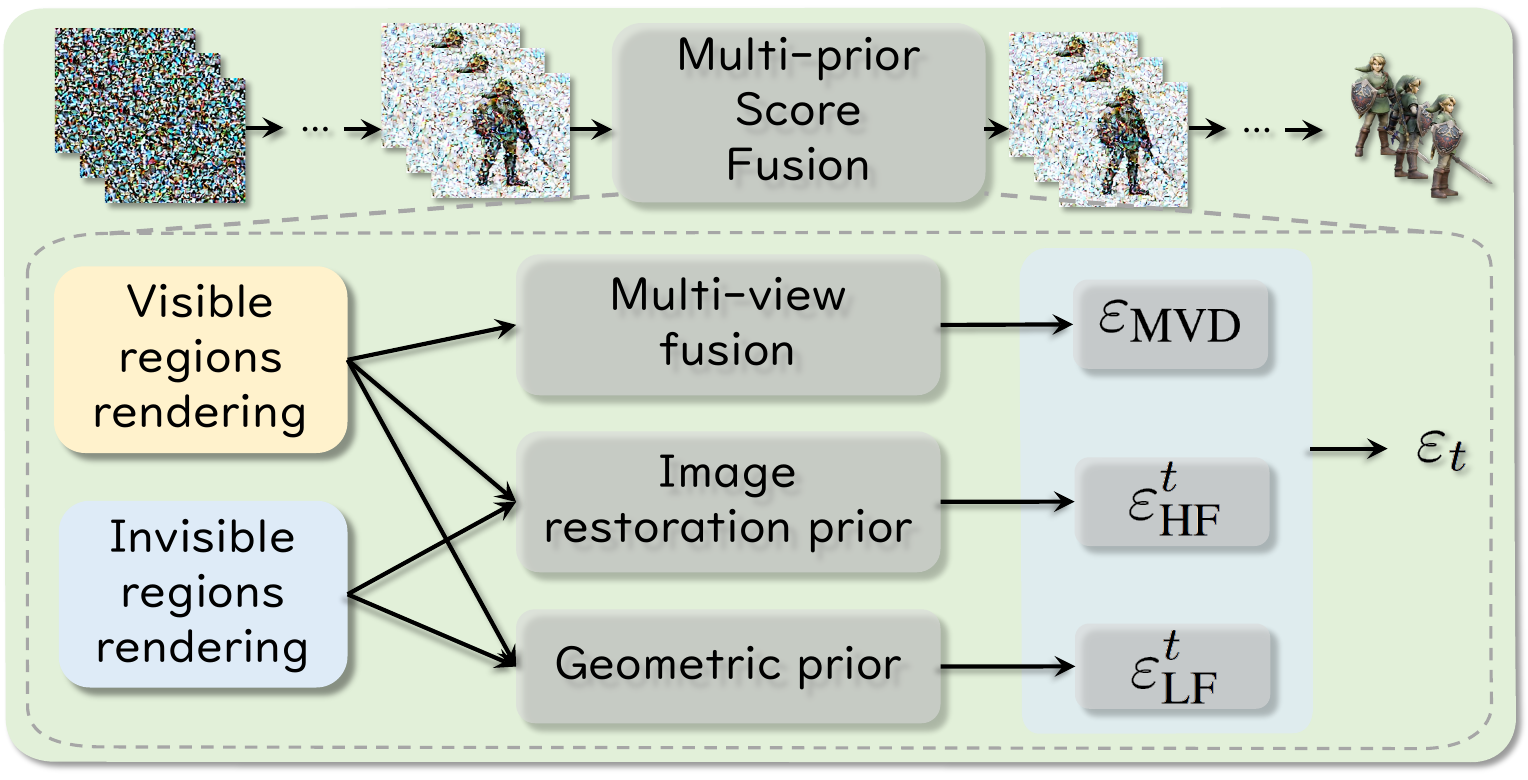}
    \caption{The Multi-Prior Score Fusion framework combines three components to generate novel view images: multi-view diffusion ($\varepsilon_{MVD}$) for fusing reference view information, geometric prior ($\varepsilon_{LF}^t$) for structural consistency, and image restoration prior ($\varepsilon_{HF}^t$) for enhancing details. These components process both visible and invisible regions from reference views and are combined with time-dependent weights to produce the noise prediction $\varepsilon_t$ used in DDIM sampling for the final image generation.}
    \vspace{-10pt} 
    \label{mvd2}
\end{figure}
The key contribution of our approach is the integration of complementary priors that operate across different frequency bands, effectively regularizing against the invalid independence assumption in multi-view fusion. We incorporate two critical regularizing components:

\noindent \textbf{(i) Geometric Rendering prior} 
Derived from the initial 3DGS model $\mathcal{G}_0$, this prior provides essential low-frequency (LF) structural regularization. It generates a coarse rendering $x_{\text{LF}}$ at the target viewpoint, with noise prediction:
\begin{equation}
\label{eq:lf_prior}
\vspace{-5pt}
\varepsilon_{\text{LF}}^t = \frac{x_t - \sqrt{\alpha_t} x_{\text{LF}}}{\sqrt{1 - \alpha_t}},
\end{equation}
where $\varepsilon_{\text{LF}}^t$ is the low-frequency noise prediction at timestep $t$, $x_t$ is the noisy image at timestep $t$, $\alpha_t$ is the noise schedule parameter, and $x_{\text{LF}}$ is the coarse rendering result.This structural prior directly addresses the limitations of the multi-view approximation by enforcing global geometric consistency independent of reference views. Rather than relying on potentially correlated view information, it provides an orthogonal, geometry-based regularization that stabilizes the generation process, particularly in regions where view correlation issues are most problematic.

\noindent \textbf{(ii) Image Restoration prior} 
This complements the geometric guidance by enhancing high-frequency (HF) features such as textures and edges. This texture-focused prior generates noise predictions $\varepsilon_{\text{HF}}^t$ that refine local details, which are often inadequately captured by either the geometric prior or the multi-view fusion. By focusing on textures and fine details, this prior further mitigates the effects of the invalid independence assumption by introducing information beyond what is captured in the potentially correlated views.

We combine these regularizing priors with the MVD framework using time-dependent weights that align with the diffusion process:
\begin{equation}
\label{eq:fused_noise}
\varepsilon_t = \varepsilon_{\text{MVD}} + w_{\text{HF}}(t) \varepsilon_{\text{HF}}^t + w_{\text{LF}}(t) \varepsilon_{\text{LF}}^t,
\end{equation}
where $\varepsilon_t$ is the combined noise prediction, $\varepsilon_{\text{MVD}}$ is the base multi-view diffusion noise prediction, $w_{\text{HF}}(t)$ and $w_{\text{LF}}(t)$ are time-dependent weights, and $\varepsilon_{\text{HF}}^t$ and $\varepsilon_{\text{LF}}^t$ are the high-frequency and low-frequency noise predictions, respectively.
The weights follow a smooth transition schedule:
\begin{equation}
\label{eq:weights}
\footnotesize
w_{\text{LF}}(t) = \frac{\tanh(-(t - \tau)/\sigma) + 1}{2}, \  w_{\text{HF}}(t) = \frac{1 - w_{\text{LF}}(t)}{\eta},
\end{equation}
where $\tau$ controls the transition point, $\sigma$ determines the smoothness, and $\eta$ scales the high-frequency weight. These parameters govern the shift from low to high-frequency regularization during diffusion, ensuring geometric structure dominates at large $t$ while texture refinement becomes prominent at small $t$.

By introducing geometric and texture priors that operate orthogonally to the multi-view conditioning, we effectively compensate for the invalid independence assumption inherent in standard multi-view fusion. The geometric prior provides structural coherence that counteracts the correlation-induced artifacts, while the texture prior enhances details that might be lost or blurred.
\subsubsection{Sampling Process}
We integrate the regularized noise prediction \(\varepsilon_t\) into the Denoising Diffusion Implicit Model (DDIM) sampling:
\vspace{-3pt}
\begin{equation}
\label{eq:ddim}
x_{t-1} = x_t + (\sqrt{\alpha_{t-1}} - \sqrt{\alpha_t}) \frac{\varepsilon_t}{\sqrt{1 - \alpha_t}}.
\end{equation}
This efficient sampling approach enables the generation of high-quality novel views, even for viewpoints significantly distant from the observed views.

As illustrated in Fig.~\ref{mvd2}, through the combination of multi-view fusion and frequency-based regularization, our multi-prior score fusion method effectively addresses the theoretical limitations arising from the invalid conditional independence assumption in conventional approaches. By providing robust geometric structure and detailed texture guidance, we can mitigate artifacts caused by view correlations, resulting in synthesized images with improved coherence and fidelity across both visible and invisible regions.

\subsection{Iterative Refinement Strategy}
\label{sec:3.3}
\begin{figure}[t]
    \centering
    \includegraphics[width=1\linewidth]{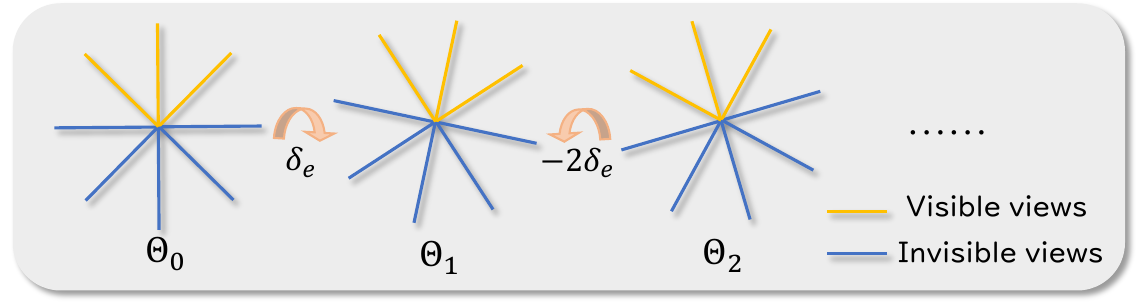}
    \caption{Schematic diagram of rotated views. Yellow lines indicate visible views used as conditional inputs for fusion-based inference, while blue lines represent invisible views that provide geometric and coarse texture information requiring restoration. Starting from initial angle $\Theta_0$, viewpoints are rotated in batches of $N$ to angles $\Theta_1$, $\Theta_2$, etc., progressively densifying the viewpoints until complete dense supervision of the object is achieved.}
    \label{fig:rotate}
    \vspace{-10pt} 
\end{figure}
The iterative refinement strategy of our pipeline employs a systematic view rotation approach to progressively enhance the 3DGS model. Starting from the initial angle $\Theta_0$ of input views, we generate batches of rotated viewpoints at incremental angles $\Theta_1, \Theta_2, \ldots$, gradually expanding coverage around the object (Fig.~\ref{fig:rotate}). For each batch of novel viewpoints, we generate high-quality, view-consistent images using our diffusion-based score fusion approach and use these images as supervision signals to refine the 3DGS model by optimizing Gaussian parameters with a composite loss function:
\begin{equation}
    \mathcal{L}_{\text{total}} = \mathcal{L}_{\text{rec}} + \lambda \mathcal{L}_{\text{LPIPS}},
\end{equation}
where $\mathcal{L}_{\text{rec}}$ measures pixel-wise reconstruction error and $\mathcal{L}_{\text{LPIPS}}$ computes the LPIPS loss using a pretrained VGG network~\cite{simonyan2014very}.
During initial iterations, we densify the Gaussian representation by adding points in under-represented regions, followed by strategic pruning of low-opacity Gaussians to maintain computational efficiency. This rotated view sampling approach ensures progressive enhancement of both geometric and appearance details across the entire object, even in regions initially unobserved.

%% file: 4_experiments.tex
\begin{figure*}
    \centering
    \includegraphics[width=1\linewidth]{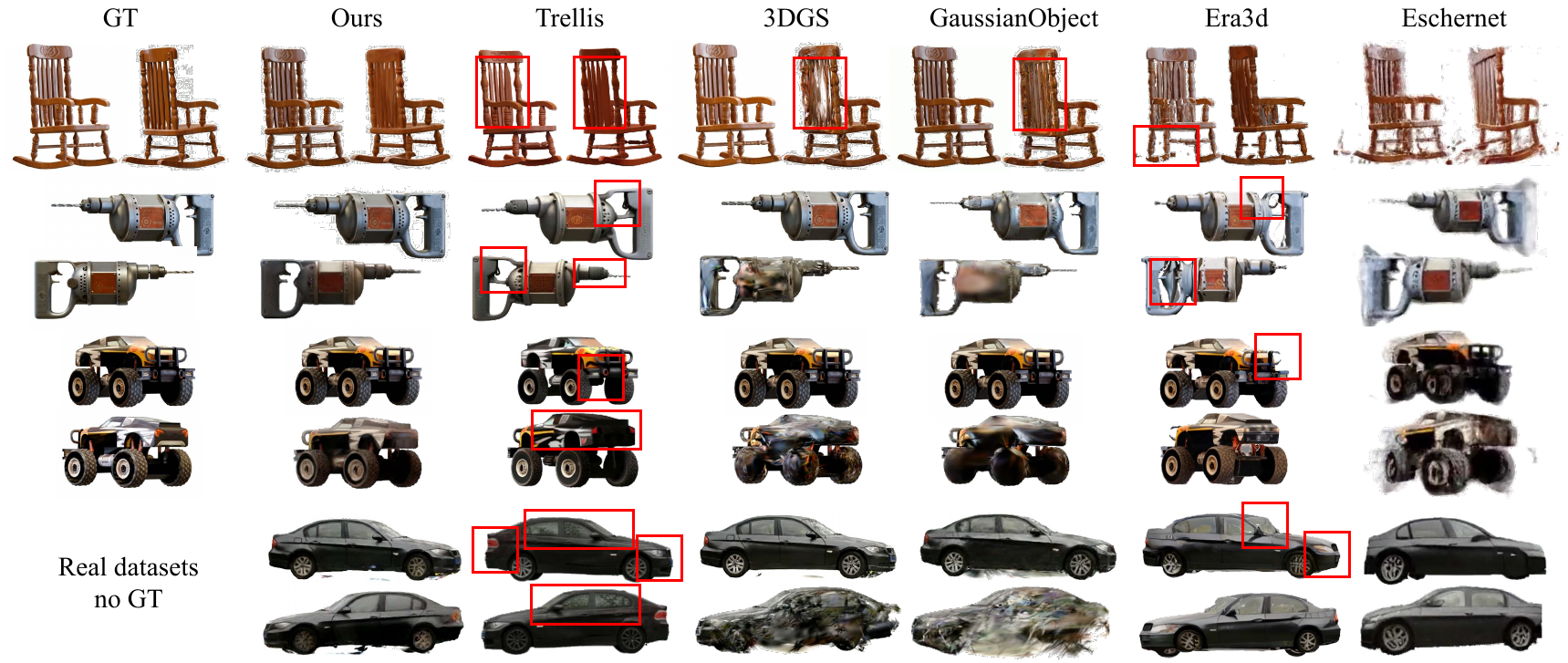}
    \caption{Qualitative comparisons between our method and other approaches. Reconstruction-based methods (3DGS, GaussianObject) achieve high fidelity in visible regions but struggle with invisible regions. In contrast, generative-based methods demonstrate the ability to synthesize reasonable geometric structures but often suffer from inconsistencies between generated content and observations (Trellis) or artifacts caused by generation inconsistencies (Era3D, EscherNet), particularly in the regions highlighted by red boxes. Our method successfully preserves details in visible regions while maintaining consistency across novel viewpoints.More qualitative results and comparisons are provided in the supplementary material.}
    \vspace{-5pt} 
    \label{fig:comparison}
\end{figure*}
\section{Experiments}
\label{sec:experiments}
\subsection{Experimental Setup}
We conduct extensive experiments on both synthetic and real-world datasets to evaluate the effectiveness of our approach. For synthetic evaluation, we utilize the Objaverse dataset~\cite{objaverseXL}, which provides ground truth images for quantitative assessment using SSIM~\cite{wang2004ssim}, LPIPS~\cite{zhang2018lpips}and PSNR metrics. Our real-world evaluation encompasses RefNeRF dataset and a custom dataset captured using an iPhone camera, where we employ the Segment Anything Model (SAM)~~\cite{kirillov2023segment} for background removal. 

Our implementation builds on the Era3D~\cite{li2024era3d} multi-view diffusion model as the MVD backbone and integrates DiffBIR as the image restoration backbone. The diffusion process employs DDIM sampling with 50 steps. During refinement, we implement a progressive rotation strategy with controlled angular offsets in alternating directions to ensure comprehensive coverage. Each refinement iteration comprises 3,000 optimization steps for the Gaussian points, with densification applied during the initial 1,300 steps. To accommodate Era3D's orthographic projection output, we adjust the rendering camera by increasing its distance by a factor of 100 and modifying the focal length accordingly. The hyper-parameters $\tau$, $\sigma$, $\eta$, $\Delta\theta$, $\delta_e$ and $\lambda$ are set to 22, 7, 4, 45, 6, and 1, respectively. 

\subsection{Main Comparison}
We compare our Zero-P-to-3 method against state-of-the-art approaches in both single-view and multi-view reconstruction categories. For a fair comparison, each method is adapted to our experimental setting: Zero-P-to-3 (Ours) uses 90 images (30 each at $0^{\circ}$, $30^{\circ}$, $-30^{\circ}$ elevation) covering $90^{\circ}$ horizontally; GaussianObject~\cite{yang2024gaussianobject} and EscherNet~\cite{kong2024eschernetgenerativemodelscalable} use 30 images at $0^{\circ}$ elevation; Trellis~\cite{xiang2024structured3dlatentsscalable} utilizes 5 images at $0^{\circ}$, $30^{\circ}$, $45^{\circ}$, $60^{\circ}$, and $90^{\circ}$ horizontal angles (all at $0^{\circ}$ elevation); Era3D~\cite{li2024era3d}, LGM~\cite{tang2025lgm}, and OpenLRM~\cite{hong2023lrm} each use a single image captured at $45^{\circ}$ horizontal, $0^{\circ}$ elevation; and 3DGS~\cite{kerbl20233d} uses the same 90 images as our method. This set of methods includes both reconstruction-based (3DGS, GaussianObject) and generative-based (Era3D, LGM, OpenLRM, EscherNet, Trellis) techniques, enabling evaluation of different paradigms for 3D reconstruction from partial observations.
\begin{table*}[ht]
\centering
\small
\begin{tabular}{c|ccc|ccc|ccc}
\toprule[1.5pt]
\textbf{Method} & \multicolumn{3}{c|}{\textbf{Visible Region}} & \multicolumn{3}{c|}{\textbf{Invisible Region}} & \multicolumn{3}{c}{\textbf{Total Region}} \\
\cmidrule(lr){2-4} \cmidrule(lr){5-7} \cmidrule(lr){8-10}
 & PSNR $\uparrow$ & SSIM $\uparrow$ & LPIPS $\downarrow$ & PSNR $\uparrow$ & SSIM $\uparrow$ & LPIPS $\downarrow$ & PSNR $\uparrow$ & SSIM $\uparrow$ & LPIPS $\downarrow$ \\
\midrule[1pt]
{3DGS} & \textcolor{red!100}{\textbf{20.64}} & \textcolor{red!100}{\textbf{0.8958}} & \textcolor{red!100}{\textbf{0.0516}} & \textcolor{pink!150}{\textbf{17.63}} & 0.8319 & 0.1222 & \textcolor{pink!150}{18.38} & \textcolor{pink!150}{\textbf{0.8479}} & \textcolor{pink!150}{0.1045} \\
{Era3d} & 16.23 & 0.8392 & 0.1312 & 15.60 & 0.8325 & 0.1399 & 15.76 & 0.8342 & 0.1377 \\
{LGM} & 12.58 & 0.7924 & 0.2078 & 12.34 & 0.7874 & 0.2179 & 12.40 & 0.7887 & 0.2154 \\
{LRM} & 13.84 & 0.8044 & 0.1896 & 12.56 & 0.8043 & 0.2098 & 12.88 & 0.8043 & 0.2048 \\
{EscherNet} & 18.68 & 0.8313 & 0.2275 & 15.61 & 0.7971 & 0.2636 & 16.38 & 0.8057 & 0.2546 \\
{GaussianObject} & 17.57 & 0.8535 & 0.0924 & 16.79 & 0.8360 & 0.1358 & 16.98 & 0.8403 & 0.1250 \\
{LGM (Era3d)} & 14.45 & 0.8145 & 0.1729 & 14.25 & 0.8154 & 0.1758 & 14.30 & 0.8152 & 0.1751 \\
{Trellis} & 16.68 & 0.8461 & 0.1121 & 16.31 & \textcolor{pink!150}{\textbf{0.8382}} & \textcolor{pink!150}{\textbf{0.1188}} & 16.40 & 0.8402 & 0.1172 \\
{\method~(Ours)} & \textcolor{pink!150}{\textbf{20.53}} & \textcolor{pink!150}{\textbf{0.8896}} & \textcolor{pink!150}{\textbf{0.0604}} & \textcolor{red!100}{\textbf{18.14}} & \textcolor{red!100}{\textbf{0.8547}} & \textcolor{red!100}{\textbf{0.1016}} & \textcolor{red!100}{\textbf{18.74}} & \textcolor{red!100}{\textbf{0.8634}} & \textcolor{red!100}{\textbf{0.0913}} \\
\bottomrule[1.5pt]
\end{tabular}

\caption{Quantitative comparison across different approaches. The table presents PSNR, SSIM, and LPIPS metrics for visible regions, invisible regions, and total reconstruction. \textcolor{red!100}{Red} numbers indicate best performance while \textcolor{pink!150}{Pink} indicates second best. 3DGS overfits to visible areas but performs poorly in invisible regions. Trellis demonstrates decent results in invisible areas but shows notable differences from ground truth. LGM (Era3D) represents results from replacing LGM's backbone with Era3D. Our Zero-P-to-3 method achieves superior overall performance, effectively balancing fidelity in visible regions with consistent generation in invisible regions.}
\vspace{-10pt}
\label{tab:quant-compare}
\end{table*}

As illustrated in Table~\ref{tab:quant-compare}, our~\method~demonstrates superior performance on the Objaverse dataset across all evaluation metrics. Reconstruction-based methods like 3DGS achieve excellent fidelity in visible regions but struggle significantly with invisible regions. In contrast, generative methods such as Era3D and Trellis show more balanced performance between visible and invisible regions, though their overall metrics fall short of 3DGS due to the accuracy in visible regions of 3DGS. Our Zero-P-to-3 successfully combines the strengths of both paradigms, maintaining high fidelity in visible regions while generating consistent and realistic content in invisible regions. Qualitative evaluation in Fig.~\ref{fig:comparison} further supports these findings, showing that our method produces more coherent geometric structures and preserves finer surface details. The areas highlighted in red boxes demonstrate where competing methods generate content that deviates significantly from ground truth, exhibiting inconsistencies with the input observations. Additional results are available in supplementary material.

\subsection{Ablation Study}
\begin{table}[ht]
\centering

\footnotesize
\begin{tabular}{c|lccc}
\toprule
No. & Method & PSNR $\uparrow$ & SSIM $\uparrow$& LPIPS $\downarrow$ \\
\midrule
& \multicolumn{4}{c}{\textbf{Based on Era3D}} \\
1 & Era3D + NeuS & 15.76 & 0.8432 & 0.1377 \\
2 & Era3D + MF + NeuS & 16.06 & 0.8542 & 0.1310 \\
\midrule
& \multicolumn{4}{c}{\textbf{Based on 3DGS}} \\
3 & 3DGS & 18.38 & 0.8479 & 0.1045 \\
4 & 3DGS + MF & 18.25 & 0.8559 & 0.0914 \\
5 & 3DGS + MF + RVS & 18.68 & 0.8615 & 0.0996 \\
6 & 3DGS + MF + RVS + IRP & 18.66 & 0.8579 & 0.0958 \\
7 & 3DGS + MF + RVS + GP & 18.38 & 0.8630 & 0.0921 \\
8 & Ours & \textbf{18.74} & \textbf{0.8634} & \textbf{0.0913} \\
\bottomrule

\end{tabular}
\caption{Ablation study on 3D scene reconstruction with diffusion models. Adding Mulit-view Fusion(MF), Rotated View Sampling (RVS), Image Restoration Prior (IRP), and Geometric Prior (GP) enhances detail and quality, with \method~ combining all to achieve the best results, showing their combined strength.}

\label{tab:ablation}
\end{table}

\begin{figure}[ht]
    \centering
    \includegraphics[width=0.9\linewidth]{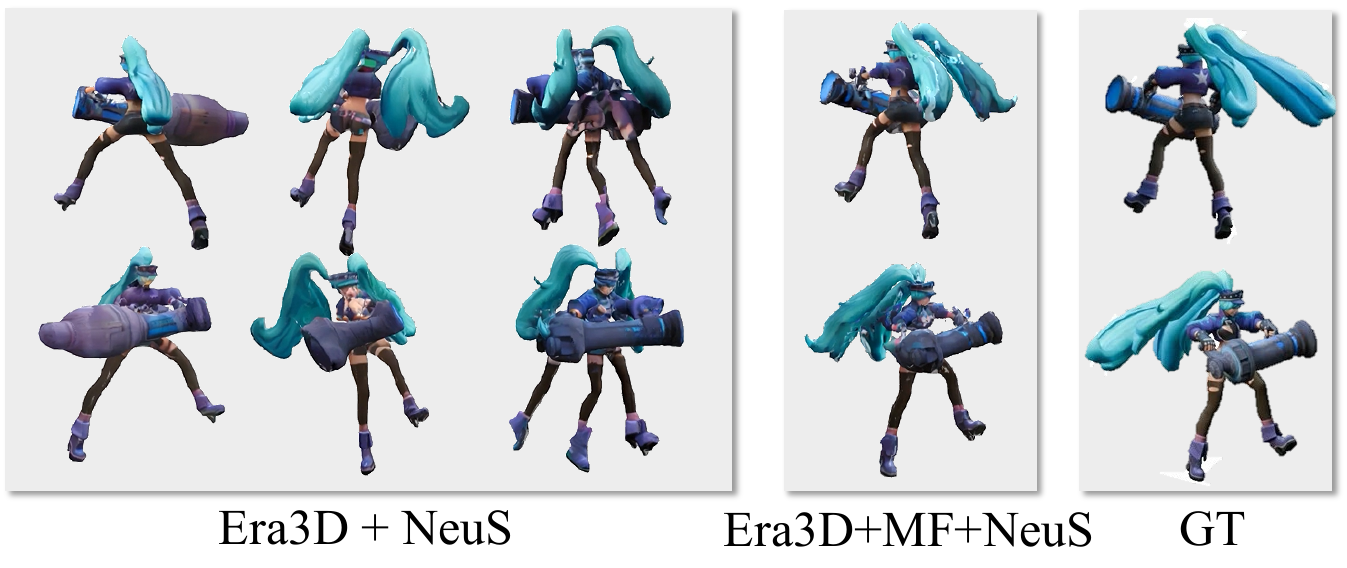}
    \caption{Ablation study on the effect of Multi-view Fusion (MF). The reconstruction using multi-view fusion with NeuS achieves high similarity to the ground truth (GT). }
    \label{fig:ablation1}
    \vspace{-6pt} 
\end{figure}

\noindent
\textbf{Analysis of Multi-view Fusion.}
As shown in Fig.~\ref{fig:ablation1}, multi-view fusion significantly mitigates the issue of missing information from single-view inputs, greatly enhancing the quality and fidelity of the generated objects. This is particularly crucial in our setting, where observations are limited to partial views. These visual results align with the quantitative data presented in rows 1 and 2 of Table~\ref{tab:ablation}.

\begin{figure}[ht]
    \centering
    \includegraphics[width=1\linewidth]{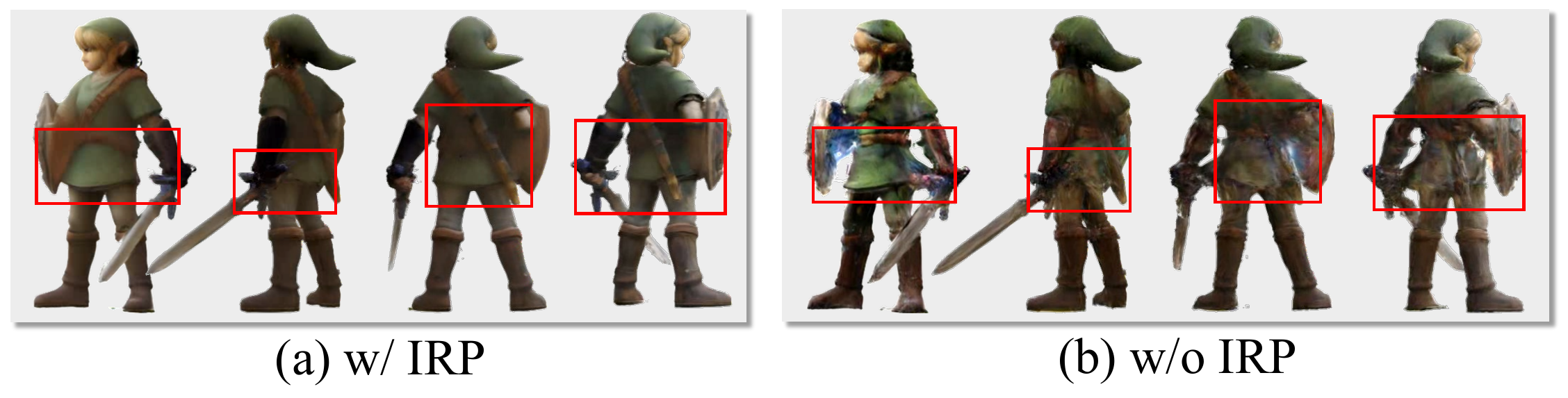}
    \caption{Effects of applying the Image Restoration Prior (IRP) in multi-view fusion. The red boxes highlight notable quality differences, where images generated without the restoration prior exhibit degraded results compared to those with the prior enabled.}
    \label{fig:ablation2}
\end{figure}

\noindent \textbf{Analysis of Image Restoration Prior.}
The integration of the image restoration diffusion model also proves crucial for detail enhancement. The quality of the leftmost image in the second row of Fig.~\ref{fig:ablation2} (b) is the worst, which is because our rotation iteration strategy inevitably takes the rendering results of invisible views as the image condition for the input of MVD. However, the inference goal of MVD for the image is to restore the image condition as much as possible, which leads to the failure of the repair. The comparison results between Fig.~\ref{fig:ablation2} (a) and Fig.~\ref{fig:ablation2} (b) show that introducing the image restoration prior has solved this problem and can improve the repair quality in more invisible views. These visual results align with the quantitative data presented in rows 7 and 8 of Table~\ref{tab:ablation}.

\begin{figure}[H]
    \centering
    \includegraphics[width=1\linewidth]{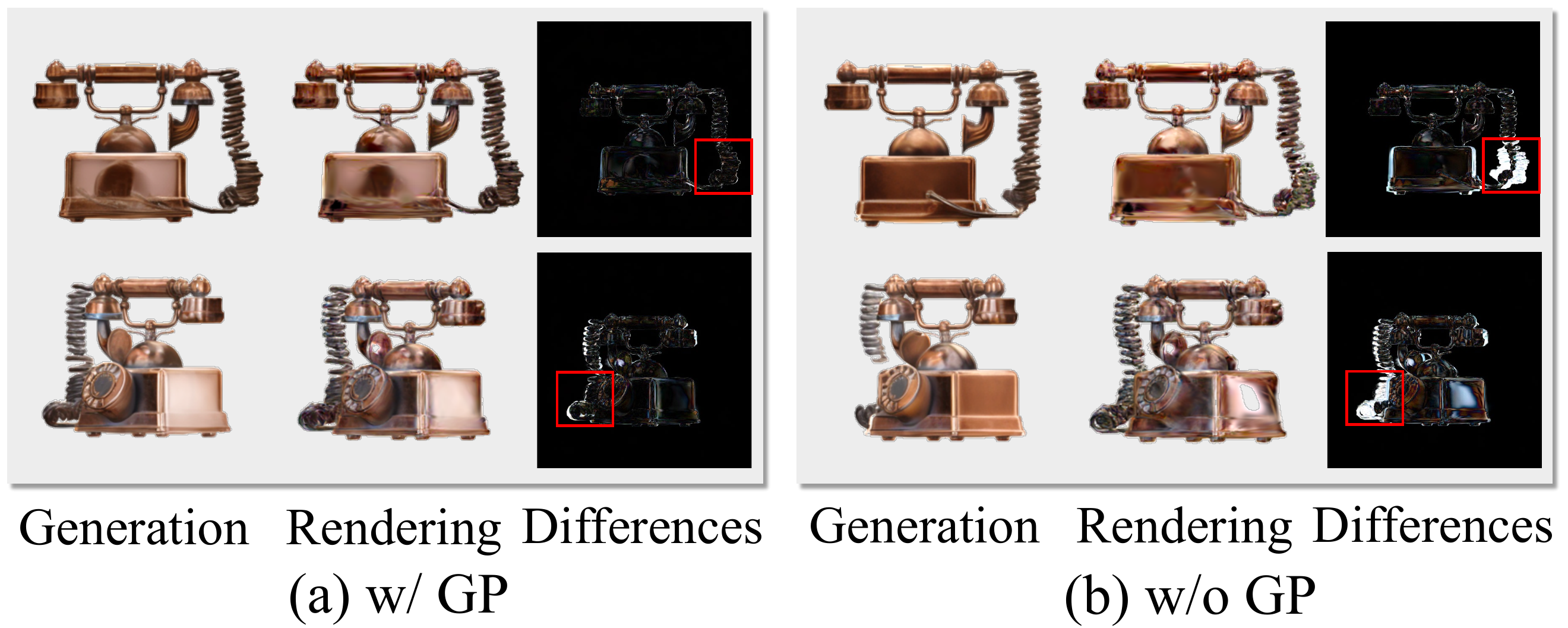}
    \caption{Effects of Geometric Prior in multi-view fusion. Each subplot presents diffusion-generated images, 3DGS-rendered images, and their differences. Red boxes in (b) (without Prior) indicate structural flaws, while (a) (with Prior) exhibits enhanced fidelity, highlighting the method's importance.}
    \label{fig:ablation3}
    \vspace{-5pt} 
\end{figure}

\noindent \textbf{Analysis of Rendering Conditioned Inference.}
As described by Eq.~\eqref{eq:lf_prior} and Eq.~\eqref{eq:fused_noise}, our \method~ incorporates the rendering results of 3DGS as the geometric prior for generating images. Figure \ref{fig:ablation3} underscores the importance of our method. Without rendering guidance, generated results exhibit significant structural discrepancies from the original image. In contrast, \method~ produces images that maintain structural consistency and high-quality textures, ensuring effective object restoration. Quantitative results in Table \ref{tab:ablation} (rows 6 and 8) further support this improvement.

\begin{figure}[t]
    \centering
    \includegraphics[width=1\linewidth]{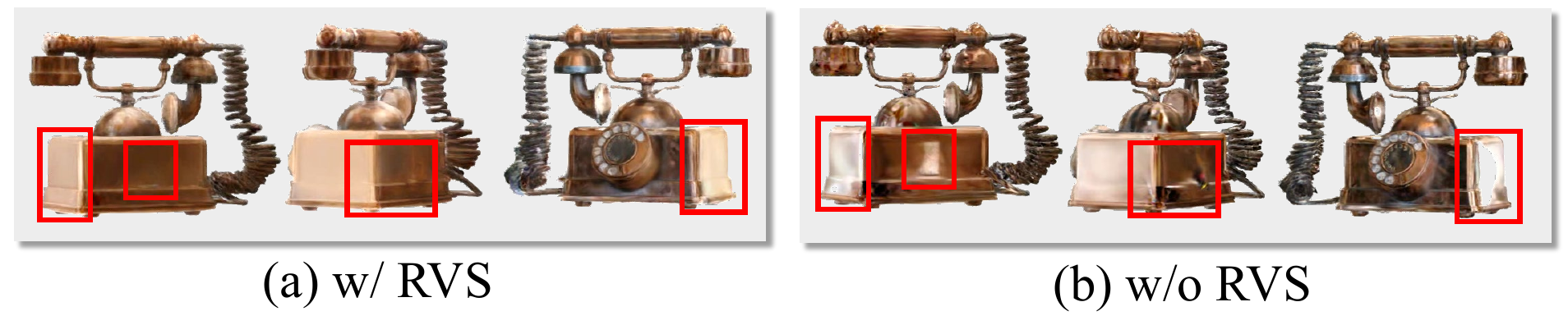}
    \caption{Effects of rotated view sampling. The regions framed in red show that there is a significant difference in texture quality between the two under specific viewing angles.}
    \label{fig:ablation5}
    \vspace{-5pt} 
\end{figure}

\noindent \textbf{Analysis of Rotated View Sampling.}
Figure~\ref{fig:ablation5} reveals that the anisotropy of 3DGS~\cite{kerbl20233d} causes artifacts when repairing objects using only a few fixed viewing angles. This highlights the critical role of our rotated view sampling method, which densifies the supervision views, significantly enhancing rendering quality from novel perspectives and ensuring high-fidelity object reconstruction.

\begin{figure}[H]
    \centering
    \includegraphics[width=1\linewidth]{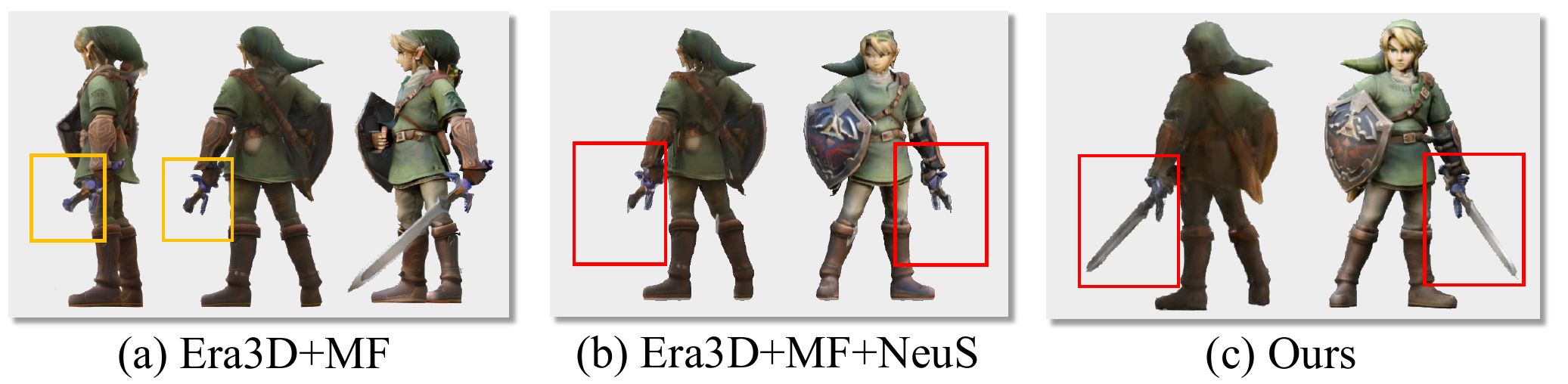}
    \caption{Analysis of direct reconstruction using results generated by Era3D. Yellow boxes highlight the failure to generate multi-view images of the sword, while red boxes indicate the failure of sword reconstruction. (a) shows the poor result after simply applying multi-view fusion, and this further leads to problems in the reconstruction results shown in (b).  }
    \label{fig:analysis}
\end{figure}

\noindent \textbf{Analysis of Iterative Refinement Strategy.} Fig.~\ref{fig:analysis} clearly demonstrates how our iterative refinement strategy and fusion-based inference method work together to fundamentally mitigate the issue caused by the conditional independence approximation. In traditional Multi-View Diffusion (MVD) inference, simply averaging the noisy predictions from different view conditions (as shown in Eq.~\eqref{eq:mvd}) can result in conflicts in regions where the information from the views does not overlap. For instance, in the sword region, different views give entirely different predictions, prevent-

\noindent ing consensus and leading to the area being treated as background noise and gradually eliminated.

To address this limitation, we introduce two complementary priors in the multi-view fusion process. Low-frequency priors leverage initial geometric renderings to provide global structural constraints, effectively stabilizing the overall 3D layout. Simultaneously, high-frequency priors focus on recovering texture details, compensating for the blurring issues caused by the lack of correlation across multi-view information. These complementary priors to-
\noindent gether resolve the issue of the invalid conditional independence assumption, enabling the iterative refinement process to utilize the original image's rendered results to guide MVD sampling at each iteration, continuously improving the quality and 3D consistency of the generated images. Experimental results (see rows 2 and 8 in Table 2) confirm the effectiveness of this strategy in resolving conflicts and recovering missing details.

%% file: 5_conclusion.tex
\section{Conclusion}
\label{sec:conclusion}
We present Zero-P-to-3, a novel training-free approach for high-quality 3D reconstruction from partial observations. Our method effectively addresses the challenges of partial-view reconstruction through a pipeline that combines multi-view diffusion priors, image restoration, and iterative refinement with rotated view supervision. Experiments show its superior performance in both reconstruction quality and multi-view consistency. The \method~ framework's success in handling incomplete observations opens doors to future work, including integration with other 3D representations and application to dynamic scenes.
